\documentclass[letterpaper, 10 pt, conference]{ieeeconf}  

\IEEEoverridecommandlockouts                              
\usepackage[utf8]{inputenc}
\usepackage[T1]{fontenc}
\usepackage{tikz,steinmetz,gensymb}
\usepackage{lineno,pifont}
\usepackage{graphicx}
\usepackage{amsmath,amsfonts}
\usepackage{float}
\usepackage{algorithmic}
\usepackage{booktabs} 
\usepackage{multirow} 
\usepackage{tabularx} 
\usepackage[table]{xcolor} 
\usepackage{algorithm}
\usepackage{array}
\usepackage[font=small]{caption}
\usepackage{subcaption}
\usepackage{textcomp}
\usepackage{stfloats}
\usepackage{url}
\usepackage{amssymb}
\usepackage{verbatim}
\usepackage{cite}
\usepackage[colorlinks=true, allcolors=black]{hyperref} 
\usepackage{cleveref}
\usepackage{academicons} 
\usepackage{orcidlink,bm}
\usepackage{xcolor}
\usepackage{pdflscape}
\usepackage[dvipsnames]{xcolor}

\renewcommand{\keywords}[1]{\textbf{\textit{Index terms---}} #1}

\def\a{{\bm a}}

 \def\r{{\bm r}}  \def\t{{\bm t}}
 \def\v{{\bm v}}

 \def\R{{\bm R}}

\def\o{{\boldsymbol \omega}}

\title{\LARGE \bf
Robotic Tele-Operation for Upper Aerodigestive Tract Microsurgery: System Design and Validation
}

\author{
Giovanni Braglia$^{1}$,  José Jair Alves Mendes Junior$^{2}$, Augusto Tetsuo Prado Inafuco$^{2}$,\\ Federico Mariano$^{1}$ and Leonardo S. Mattos$^{1}$
\thanks{This research was partially funded by the European Union - Horizon Europe project “AIRCARE - AI-augmented Robotics for CAncer point of caRE" (101137426). 
Views and opinions expressed are however those of the authors only and do not necessarily reflect those of the European Union. Neither the European Union nor the granting authority can be held responsible for them.}
\thanks{$^{1}$ Biomedical Robotics Laboratory, Advanced Robotics, Istituto Italiano di
Tecnologia, Genoa, Italy}%
\thanks{$^{2}$ Graduate Program in Electrical and Computer Engineering (CPGEI), Federal University of Techonology - Paraná, Curitiba, Brazil.}%
\thanks{\noindent Video: \textit{https://youtu.be/dXg9meC2AcI}}
}

\begin{document}

\maketitle
\thispagestyle{empty}
\pagestyle{empty}

\begin{abstract}

Upper aerodigestive tract (UADT) treatments frequently employ transoral laser microsurgery (TLM) for procedures such as the removal of tumors or polyps. In TLM, a laser beam is used to cut target tissue, while forceps are employed to grasp, manipulate, and stabilize tissue within the UADT. Although TLM systems may rely on different technologies and interfaces, forceps manipulation is still predominantly performed manually, introducing limitations in ergonomics, precision, and controllability.
This paper proposes a novel robotic system for tissue manipulation in UADT procedures, based on a novel end-effector designed for forceps control. The system is integrated within a teleoperation framework that employs a robotic manipulator with a programmed remote center of motion (RCM), enabling precise and constrained instrument motion while improving surgeon ergonomics. The proposed approach is validated through two experimental studies and a dedicated usability evaluation, demonstrating its effectiveness and suitability for UADT surgical applications.

\end{abstract}
\keywords{ Minimally invasive surgery, Teleoperation, Robot assistance. }

\section{Introduction}\label{sec:Introduction}

Telerobotic surgery enables the transfer of a surgeon’s sensorimotor skills to a robotic platform, allowing surgical interventions to be performed on the patient’s body through a leader-follower control architecture. On one hand, by decoupling the surgeon from the operative site, telesurgery has marked a significant step forward in surgical technology, enabling procedures to be carried out remotely with enhanced accuracy and precision \cite{mohan2021telesurgery,acemoglu2020}. On the other hand, the integration of advanced sensing and haptic feedback within telerobotic systems further improves the surgeon’s sensory awareness and motor control, contributing to increased precision during manipulation and, ultimately, to improved surgical procedures and patient outcomes \cite{patel2022haptic}.

The benefits of robot-assisted telesurgery are increasingly recognized in healthcare, leading to its adoption in a growing number of surgical procedures; however, several application domains remain unexplored\cite{barba2022remote}.
Upper aerodigestive tract (UADT) diseases fall within this category. Tumors affecting this anatomical region represent, for instance, the sixth most common type of cancer and show a strong prevalence among smokers. In recent years, UADT conditions have attracted increasing research attention for several reasons. First, advances in medical imaging and artificial intelligence have enabled improved diagnostic capabilities, including tumor segmentation \cite{jiang2022deep} and analysis \cite{petersen2025development}. Second, surgical treatments have become more accurate and less invasive due to techniques such as transoral laser microsurgery (TLM) \cite{pacheco2024automatic,carrieri2025design}.

In TLM, the surgeon employs a laser micromanipulator to perform precise laser ablation for the resection of pathological tissue, such as tumor regions\cite{lee2022end}. This approach benefits patients by enabling minimally invasive and highly accurate cuts, resulting in reduced trauma and faster postoperative recovery \cite{ansarin2014transoral,carrieri2025design}. While tissue ablation is executed by the laser micromanipulator, tissue manipulation and removal are still performed manually using forceps.
\begin{figure}[t]
    \centering
    \includegraphics[trim={6cm 0.5cm 6cm 0cm},clip,width=0.96\linewidth]{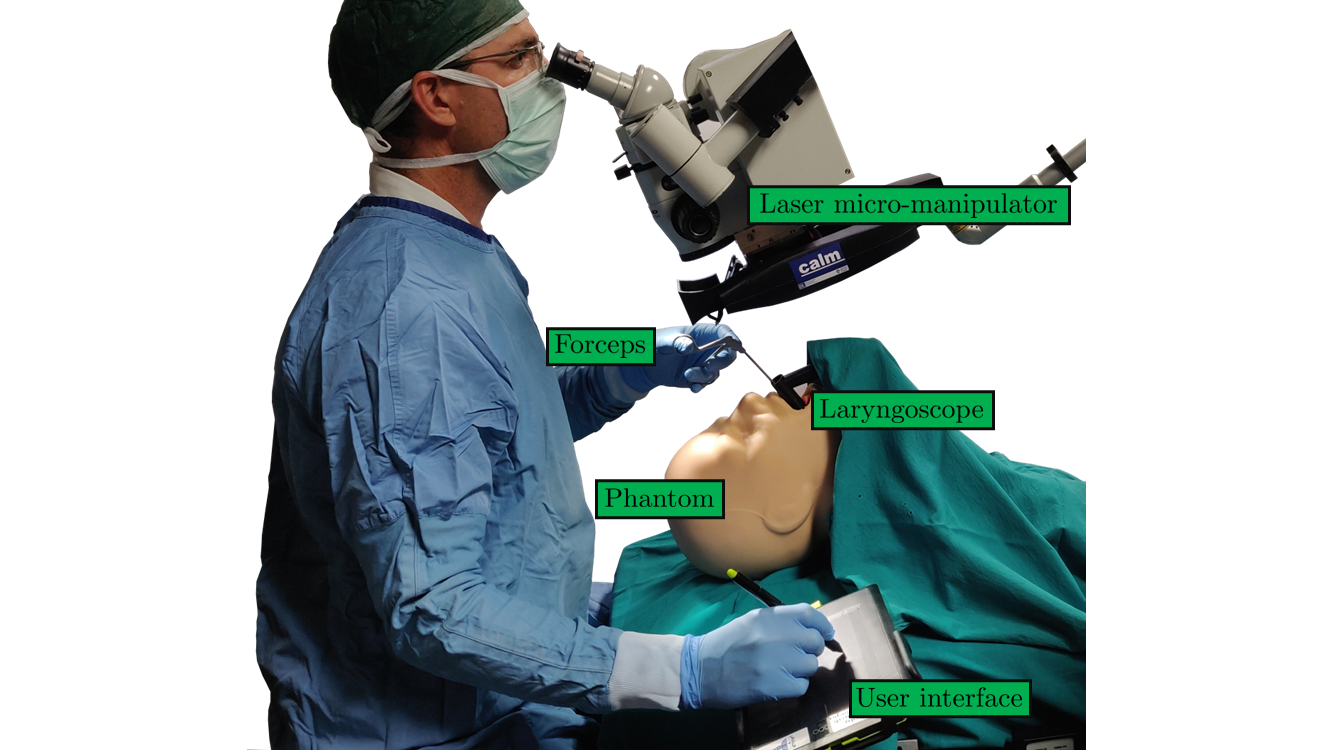}
    \caption{OR setup example during UADT procedures.}
    \label{fig:or_setup}
    \vspace{-5mm}
\end{figure}
The typical operating room (OR) setup for TLM is schematically illustrated in Fig.~\ref{fig:or_setup}. During the procedure, the patient’s mouth is kept open using a laryngoscope, which is inserted into the oral cavity and mechanically fixed to the operating table. A surgical endoscope camera is employed to provide a magnified view of the UADT, while the surgeon operates the laser micromanipulator through a dedicated user interface to perform tissue ablation \cite{carrieri2025design}.
Prior to laser ablation, the surgeon inserts forceps through the laryngoscope to grasp and manipulate the target tissue, improving visibility and positioning of the area of interest. As shown in Fig.~\ref{fig:or_setup}, this manipulation is currently performed manually, which may introduce limitations such as hand tremor, grasp instability, and surgeon fatigue, especially during prolonged or repeated interventions\cite{o2010methodology}.

Motivated by recent advances in telesurgery, this work proposes a novel teleoperated system to replace manual tissue manipulation in transoral microsurgery. Previous approaches targeting this application have been reported in \cite{chauhan2019robotic,deshpande2016robot,nasiri2024teleoperation,colan2020cooperative}. Differently from these works, the proposed system combines custom end-effectors with off-the-shelf medical-grade robotic platforms, supporting both local and remote UADT microsurgery \cite{acemoglu2020} while reducing hardware complexity and facilitating clinical translation. Moreover, the system advances the state of the art by introducing additional capabilities, including the software enforcement of a remote center of motion (RCM), similarly to \cite{aghakhani2013task}, to facilitate forceps manipulation within the laryngoscope. Overall, the proposed approach aims to improve manipulation stability and reduce surgeon effort while remaining fully compatible with the operational and spatial requirements of TLM.

Building on these design objectives, the main contributions of this work are threefold: (i) the design of a novel computer-controlled end-effector for forceps manipulation; (ii) the development of a complete teleoperation pipeline for robotic control based on a remote center of motion (RCM) constraint; and (iii) an extensive experimental validation, including a pilot user study involving the manipulation of a silicone vocal fold.
The remainder of the paper is organized as follows. Section~\ref{sec:Materials} describes the components of the proposed framework. Section~\ref{sec:Evaluation} details the experimental setup and presents the key results. Section~\ref{sec:Discussion} discusses the findings and concludes the paper.

\section{Materials and Methods}\label{sec:Materials}

%
\begin{figure*}[t]
    \centering
    \includegraphics[trim={0cm 3cm 0.5cm 3cm},clip,width=0.96\linewidth]{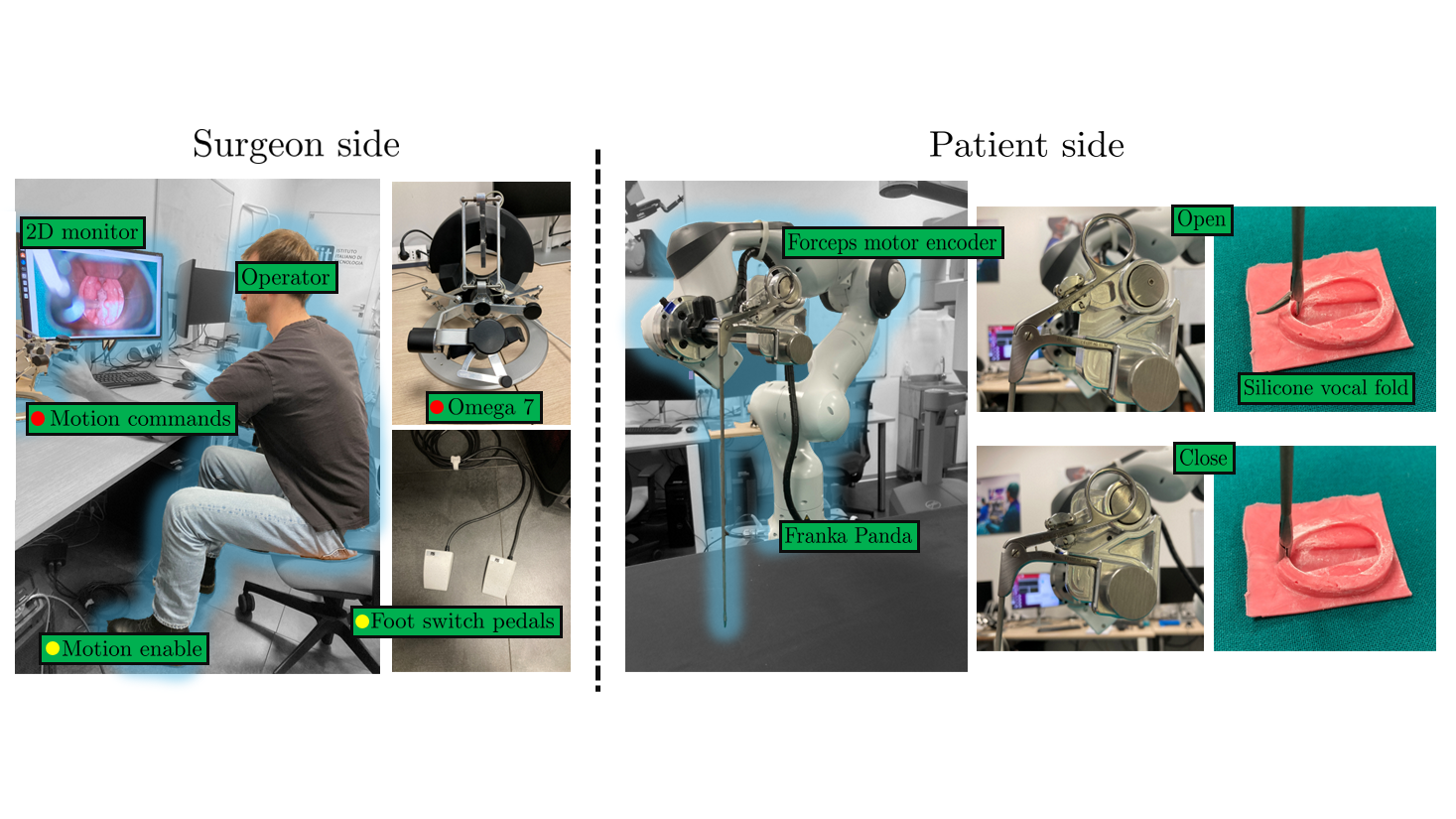}
    \caption{Proposed teleoperation system. On the surgeon side, the leader operator takes control by simultaneously pressing two enabling pedals, thus sending velocity commands with the remote controller. On the patient side, the follower robot moves our custom end-effector, whose open-close movements are controlled directly from the operator.}
    \label{fig:teleoperation_system}
    \vspace{-5mm}
\end{figure*}

The description of the proposed telesurgery system for UADT treatments is depicted in Fig.\ref{fig:teleoperation_system}. The system consists of a leader (surgeon) and a follower (patient) side. On the surgeon side, the operator sit on a chair where he/she can visualize the operating site through a 2D monitor displaying the view of an exoscopic camera located at the patient side. Visualization of the target tissue is performed through the laryngoscope, providing the surgeon a magnified view of the UADT\cite{ansarin2014transoral}.

On the patient side, the proposed forceps control end-effector is mounted on a robot manipulator. The robot reads movements commands from the surgeon side, sent through an haptic device located on the surgeon desk. Commands are read by the robot only if two safety foot switch are pressed simultaneously. Finally, the haptic device also allows to send open/close commands, which are read and processed by the end-effector controller to open/close the forceps jaws.
In the next paragraphs, the architecture and control aspects are detailed. The reader is also encouraged to have a look at the paper's video to have a comprehensive view of how the proposed system works.

\subsection{System setup}\label{subsec:system_setup}

The designed forceps control end-effector is shown on the right side of Fig.~\ref{fig:teleoperation_system}. The tool consists of an aluminum structure approximately 20~cm in length and weighing 2.1~kg, mounted on a Franka Emika Panda robot. The distal end of the aluminum structure is designed to accommodate a Karl Storz surgical forceps, which can be easily mounted or replaced using an aluminum clamp secured by a hand screw.
One handle of the forceps is rigidly fixed to ensure stable positioning, while the other is inserted in a dedicated interlocking mechanism. This interlocking component is mounted on an aluminum bracket actuated by a Maxon motor, which controls the opening and closing motion of the forceps jaws.

A Maxon EPOS4 position controller is used to control the motor driving the forceps, with a maximum output torque of 0.5~Nm. The EPOS4 module is compatible with the Robot Operating System (ROS), which is employed to manage communication among all system components. In particular, the EPOS4 controller receives open/close commands generated by the force gripper of a Force Dimension Omega.7 device located at the surgeon side.
The Omega.7 is a 7-DoF (6-DoF for motion + 1 DoF for the gripper) haptic device with active gravity compensation and serves two main functions within the proposed framework. First, it provides open and close commands through hand grasping motions acting on its force gripper, which directly control the forceps actuation. Second, it generates Cartesian twist commands that are sent to the Franka Emika Panda robot to control the position and orientation of the forceps.

The motion of the Franka Emika Panda robot is commanded through its dedicated ROS control node and supervised by a dual foot-switch interface. This interface acts as a safety layer: robot motion is enabled only when both foot pedals are simultaneously pressed; otherwise, the robot remains stationary. Two RS foot switches are used and connected to an Arduino Nano, which publishes a binary signal indicating the pedal state (``0'' if at least one pedal is released, ``1'' if both pedals are pressed).
The complete control pipeline, including all ROS nodes, runs on a single workstation equipped with Ubuntu Linux. To support proper gravity compensation, the designed end-effector is described by a dedicated \texttt{.json} configuration file loaded into the Panda Desk interface. In addition, a full 3D model of the end-effector with its relative \texttt{.urdf} file is provided and integrated into \texttt{RViz}, enabling real-time visualization of the robot and end-effector motion.

\subsection{RCM with Velocity Control}\label{subsec:rcm}

TLM is performed within a highly confined workspace defined by the laryngoscope aperture. Visualization of the surgical site relies on a high-magnification exoscope, making the scene particularly sensitive to small instrument motions, which may easily cause visual occlusions, as reported in previous studies \cite{chauhan2019robotic,deshpande2016robot}.
By enforcing a remote center of motion (RCM) constraint at the laryngoscope entrance, translational movements at the surgeon side are mapped into pivoting motions of the forceps about the RCM, thus selecting the RCM outside the endoscope field of view significantly reduces occlusion events. The next paragraphs describe how this the RCM is implemented in software within the proposed architecture.

The controller receives user commands in the form of a Cartesian twist $\bf{v}^{in} $$= [ \v^{in}, \o^{in} ]^\top \in \mathbb{R}^6$, directly acquired from the Omega.7 device. The objective of the controller is to map this input twist, expressed in the Omega.7 reference frame, into a twist $\bf{v}^{C}=$$ [ \v^C, \o^{C} ]^\top \in \mathbb{R}^6$ defined in a desired application frame~$\{C\}$, while enforcing a RCM constraint.
As a first step, the input linear and angular velocity components are rotated and scaled according to
\begin{subequations}\label{eq:voc_1}
\begin{align}
\mathbf{v}^{C} &= \alpha_t \mathbf{R}_{in}^{C} \mathbf{v}^{in}, \label{subeq:vc_1} \\
\boldsymbol{\omega}^{C} &= \alpha_r \mathbf{R}_{in}^{C} \boldsymbol{\omega}^{in}, \label{subeq:oc_1}
\end{align}
\end{subequations}
where $\alpha_t$ and $\alpha_r$ are translational and rotational scaling factors, respectively, and $\mathbf{R}_{in}^{C} \in \mathbb{R}^{3 \times 3}$ is a rotation matrix that maps the input twist into frame~$\{C\}$.\footnote{In the proposed system $\mathbf{R}_{in}^{C}$ was selected empirically to align the operator hand motions with the visual feedback provided on the 2D monitor.}

\begin{figure}[t]
    \centering
    \includegraphics[trim={6cm 5.5cm 16cm 2.5cm},clip,width=0.96\linewidth]{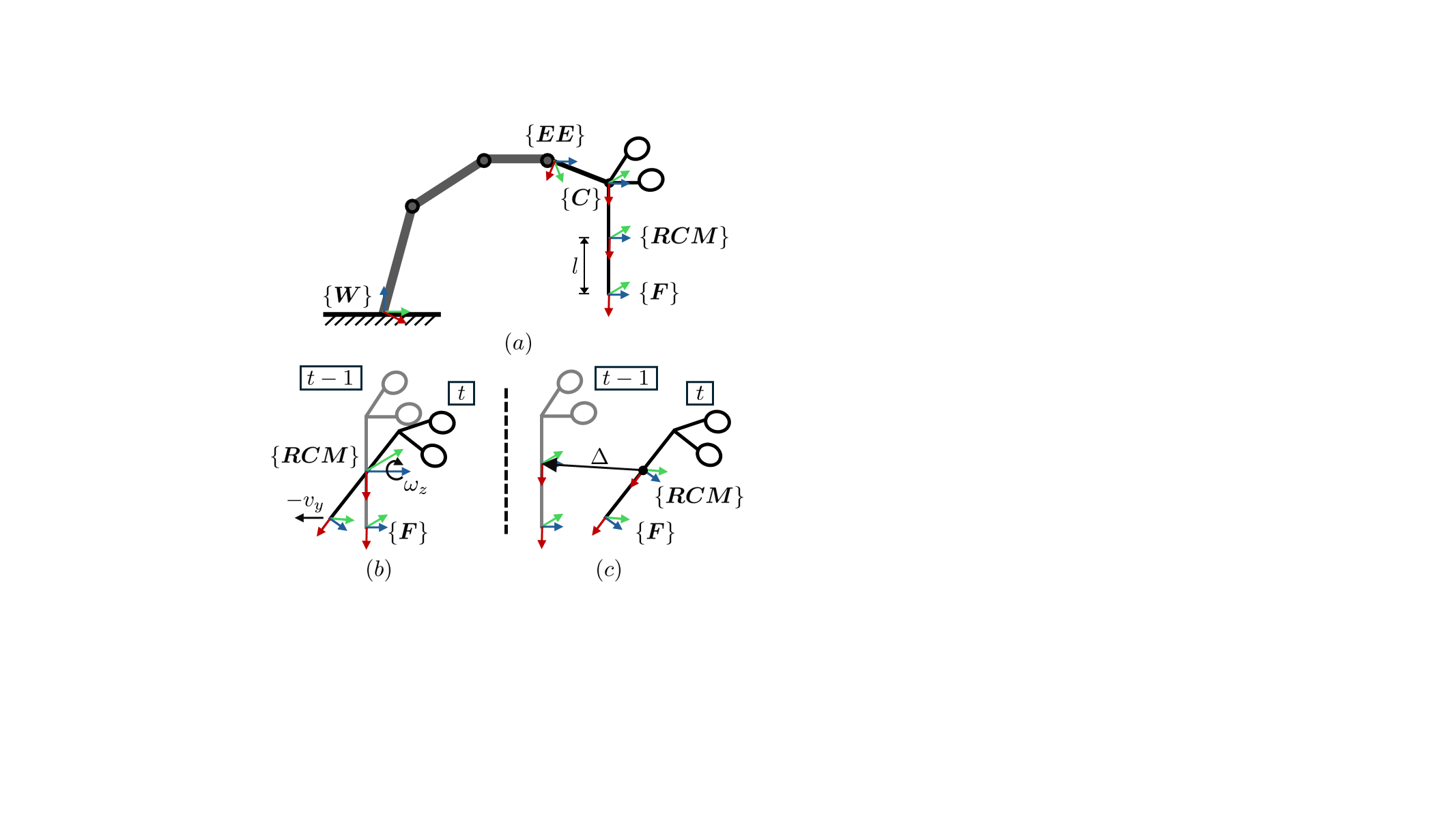}
    \caption{Schematic representation of the forceps' manipulator.}
    \label{fig:frame_scheme}
    \vspace{-5mm}
\end{figure}

At this point, the twist $\mathbf{v}^{C}$ must be processed to reproduce the user commands while enforcing a RCM constraint. This requires: (i) generating pivoting motions about the RCM, and (ii) compensating translational components to maintain the RCM fixed and prevent lateral displacements.
Within the \texttt{RViz} interface, the user specifies the desired RCM location along the forceps shaft, which defines the corresponding reference frame~$\{RCM\}$. The interface also automatically computes the reference frame associated with the forceps tip, denoted as~$\{F\}$. Both $\{RCM\}$ and $\{F\}$ are always defined with their origins located on the tool shaft and with the $x$-axis aligned with the shaft direction, as illustrated in Fig.\ref{fig:frame_scheme}(a).

The pivoting motion about the RCM is obtained by transforming translational velocity commands expressed at the forceps tip into angular velocities in the application frame~$\{C\}$. This approach exploits the fact that lateral motions observed at the tip in the 2D endoscopic view can be generated through rotations about the RCM. Figure \ref{fig:frame_scheme}(b) provides a schematic representation of this step.
Let $\r = [l,0,0]^\top$ denote the vector from the reference frame~$\{RCM\}$ to the forceps tip frame~$\{F\}$. The translational velocity $\v$ of the tip induced by an angular velocity $\o$ about the RCM is given by
\begin{equation}\label{eq:ang_vel_rcm}
    \v = \o \times \r =  \left[
    \begin{array}{c}
         0 \\
         \omega_z \cdot l   \\ 
         -\omega_y \cdot l
    \end{array}
    \right],
\end{equation}
which shows that desired lateral velocities $(v_y,v_z)$ at the forceps tip can be generated by selecting $\omega_y = -v_z/l$ and $\omega_z = v_y/l$. 
By explicitly accounting for the involved reference frames, this mapping leads to
\begin{equation}\label{eq:omega_final}
    \o^F =  \left[
    \begin{array}{c}
         \omega_x^C \\
         -v_{C,z}^F/l   \\ 
         v_{C,y}^F/l 
    \end{array}
    \right] \; \mbox{then} \; \o^C =  \R^C_F \o^F.
\end{equation}
In~\eqref{eq:omega_final}, the roll component $\omega_x^C$ remains unchanged, since the frames are aligned along the tool shaft by design. As a result, $\omega_x^F = \omega_x^{RCM} = \omega_x^C$.

To prevent the frame~$\{RCM\}$ from drifting during motion, the translational component of the twist $\v^C$ is replaced by a constrained translation through the introduction of a corrective term. Figure \ref{fig:frame_scheme}(c) provides a schematic representation of this step. Expressed in frame~$\{F\}$, the translational velocity $\v^C$ becomes
\begin{equation}\label{eq:vc_2}
    \v^F = (\R^{C}_F)^\top \v^C, 
\end{equation}
where $\R^{C}_F$ denotes the rotation matrix from frame~$\{F\}$ to frame~$\{C\}$, therefore its transpose is taken in \eqref{eq:vc_2} .


Then, the vectors $\t^{F}_{RCM}$ and $\t^{F}_{C} \in \mathbb{R}^3$ can be measured, allowing the evaluation of their difference $\Delta = \t^{F}_{RCM} - \t^{F}_{C}$. This quantity is used as a corrective term, scaled by a proportional gain $k$, leading to
\begin{equation}\label{eq:vc_rcm}
    \v^F = \left[
    \begin{array}{c}
         v^{F}_{x} \\
         k\Delta_y   \\ 
         k\Delta_z
    \end{array}
    \right], \; \mbox{then} \; \v^C =  \R^C_F \v^F,  
\end{equation}
where the component $v^{F}_{x}$ is left unchanged, as the application frame is aligned with the RCM along the $x$-axis. The final transformation expresses the corrected translational velocity in frame~$\{C\}$.
As a result, \eqref{eq:vc_rcm} computes a translational velocity $\v^C$ that compensates for deviations and maintains the RCM at its nominal position during motion.

Equations~\eqref{eq:omega_final} and~\eqref{eq:vc_rcm} are combined to compute the RCM-constrained twist
$\mathbf{v}^{C} = [\v^C, \o^{C}]^\top$. Since the Franka velocity controller requires twist commands expressed in the end-effector frame~$\{EE\}$, the final step consists in transforming $\mathbf{v}^{C}$ into~$\{EE\}$. This transformation is performed as
\begin{subequations}\label{eq:final_twist_EE}
\begin{equation}\label{subeq:final_o}
  \o^{EE} = \R_{C}^{EE} \o^{C},  
\end{equation}    
\begin{equation}\label{subeq:final_v}
  \v^{EE} = \R_{C}^{EE} \v^{C} + \t\times\o^{EE},    
\end{equation}
\end{subequations}
where $\t$ denotes the translation vector between the origins of frames~$\{C\}$ and~$\{EE\}$, expressed in the end-effector frame.

\begin{figure*}[t]
    \centering
    \includegraphics[trim={1cm 5.7cm 3cm 4cm},clip,width=0.96\linewidth]{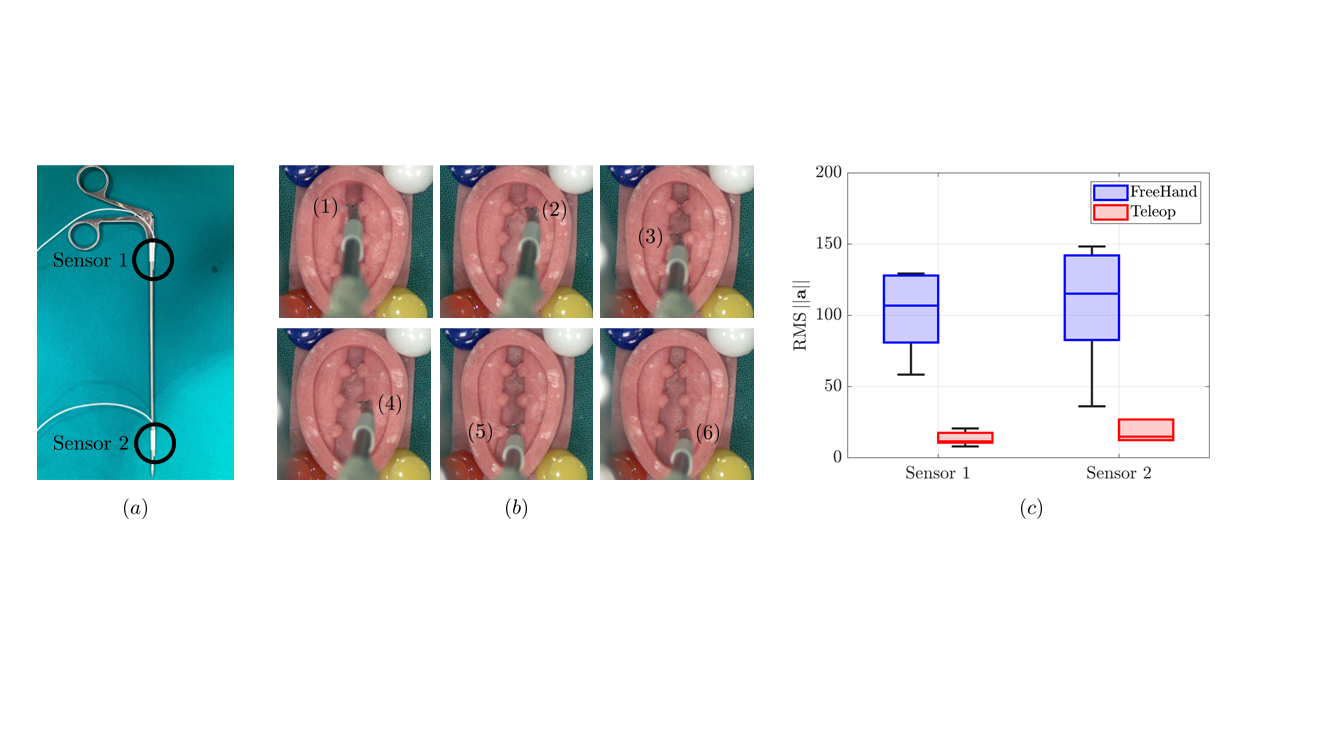}
    \caption{Controller analysis: (a) placement of the position sensors, (b) task requirement on silicone vocal fold with simulated polyps, (c) outcomes of freehand vs. teleoperation in terms of RMS of the acceleration's norm.}
    \label{fig:aurora_exp}
    \vspace{-5mm}
\end{figure*}
%

\section{Evaluation and Results}\label{sec:Evaluation}

As the proposed system is currently at an early development stage, a preliminary validation was conducted to assess its usability and potential benefits for TLM.
This validation was performed by comparing the conventional freehand forceps manipulation with the proposed teleoperation framework. Specifically, Section~\ref{subsec:controller} evaluates the handling stability of the proposed controller, Section~\ref{subsec:sEMG} analyzes the effort impact on users, and Section~\ref{subsec:usability} assesses the overall teleoperation framework from a user-centered perspective.

Ten participants with no prior TLM experience took part in the experiments. Each participant received a brief task explanation and a five-minute training session. Written informed consent to the treatment of data was obtained in accordance with EU's GDPR and ethical guidelines. This study was approved by the Ethics Committee of Liguria (Italy) with register number 229/2019 - ID 4621. A visualization of the experiments is shown in the attached video.

\subsection{Controller analysis}\label{subsec:controller}

This experiment evaluates forceps handling stability by comparing freehand manipulation with the proposed teleoperation setup. Tool motion was tracked using an NDI Aurora system, which provided real-time position measurements from two sensors mounted on the forceps, see Fig.~\ref{fig:aurora_exp}(a).

The task is illustrated in Fig.~\ref{fig:aurora_exp}(b), where a silicone vocal fold phantom with six protrusions simulating polyps was used. Participants were instructed to sequentially grasp each protrusion, maintain a stable grasp, and then gently release it. In this experiment, unlike the setup shown in Fig.~\ref{fig:or_setup}, no laryngoscope was used; therefore, the forceps motion was not constrained by an insertion channel.

Grasping stability was quantified by computing the root-mean-square (RMS) of the acceleration norm, $\mbox{RMS}\,||\a||$, measured by the two sensors. This metric provides an indication of vibration levels during forceps handling. The aggregated results are reported in the boxplots of Fig.~\ref{fig:aurora_exp}(c), which show a substantial reduction in vibrations when using the proposed teleoperation system. This improvement reflects increased grasping stability, achieved through hand tremor attenuation via velocity scaling and through the ability to immobilize the robotic arm during grasping.

%
\begin{figure}[t]
    \centering
    \includegraphics[trim={11cm 2.7cm 11cm 0.5cm},clip,width=0.96\linewidth]{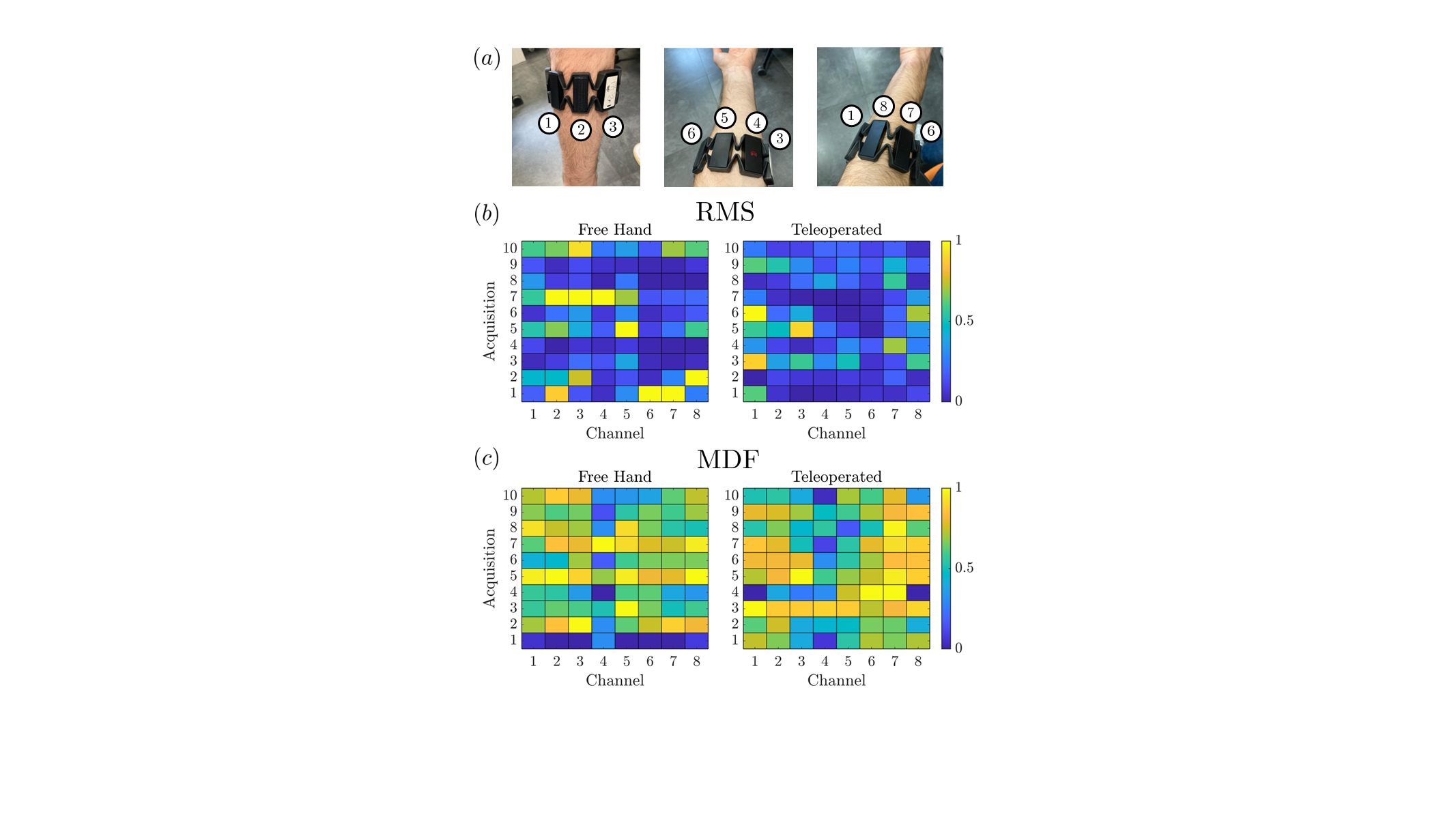}
    \caption{Outcomes of sEMG experiment.}
    \label{fig:sEMG_exp}
    \vspace{-5mm}
\end{figure}
%
\subsection{sEMG study on user effort}\label{subsec:sEMG}



The experiment described in the previous section was repeated under more realistic conditions by requiring participants to manipulate the silicone vocal fold phantom through a laryngoscope while relying on a 2D monitor for visualization, rather than looking directly at the surgical site. User effort was assessed using surface electromyography (sEMG), acquired via a Myo armband placed on the forearm used by each participant, as shown in Fig.~\ref{fig:sEMG_exp}(a). sEMG data were collected from ten participants in both freehand and teleoperated conditions to quantify muscle activation levels during task execution.

After acquisition, root-mean-square (RMS) and median frequency (MDF) features were extracted from 500~ms segments for each of the eight Myo channels. RMS was used as an indicator of muscle activation level \cite{ebied2020upper}, while MDF was employed as an inverse measure of muscle fatigue, with higher values indicating lower fatigue \cite{hou2021effects}. For each acquisition, feature values were averaged and normalized per channel. The resulting RMS and MDF outcomes are reported in Fig.~\ref{fig:sEMG_exp}(b) and Fig.~\ref{fig:sEMG_exp}(c), respectively.

During forceps manipulation, the primary muscle groups involved are the finger flexors and extensors, including the flexor pollicis longus and the finger flexors (flexor digitorum superficialis and flexor digitorum profundus). In the experimental setup, the sEMG armband was positioned such that channels 2, 3, 4, and 5 predominantly captured activity from the flexor carpi ulnaris, flexor digitorum superficialis, and flexor carpi radialis. Conversely, channels 1 and 8 were located over the extensor carpi radialis brevis and longus, which are less directly involved in the grasping task and therefore exhibited greater variability.

Although outliers are present in the RMS results for certain channels (most notably channel 1 in the teleoperation condition), this behavior is expected given the stochastic nature of sEMG signals and the inter-subject variability inherent to muscle activation patterns. Importantly, the muscle groups most relevant to forceps manipulation (channels 2–5) consistently exhibited lower median RMS values during teleoperation compared to freehand manipulation, indicating reduced muscle activation levels.
Furthermore, the median frequency (MDF) analysis showed generally higher frequency values in the teleoperation condition. Since a shift toward lower frequencies is commonly associated with muscle fatigue, the observed increase in MDF suggests reduced fatigue and lower muscular effort when using the teleoperated system.

\subsection{Usability study}\label{subsec:usability}

%
\begin{figure*}[t]
    \centering
    \includegraphics[trim={1.5cm 2.5cm 2cm 1cm},clip,width=0.96\linewidth]{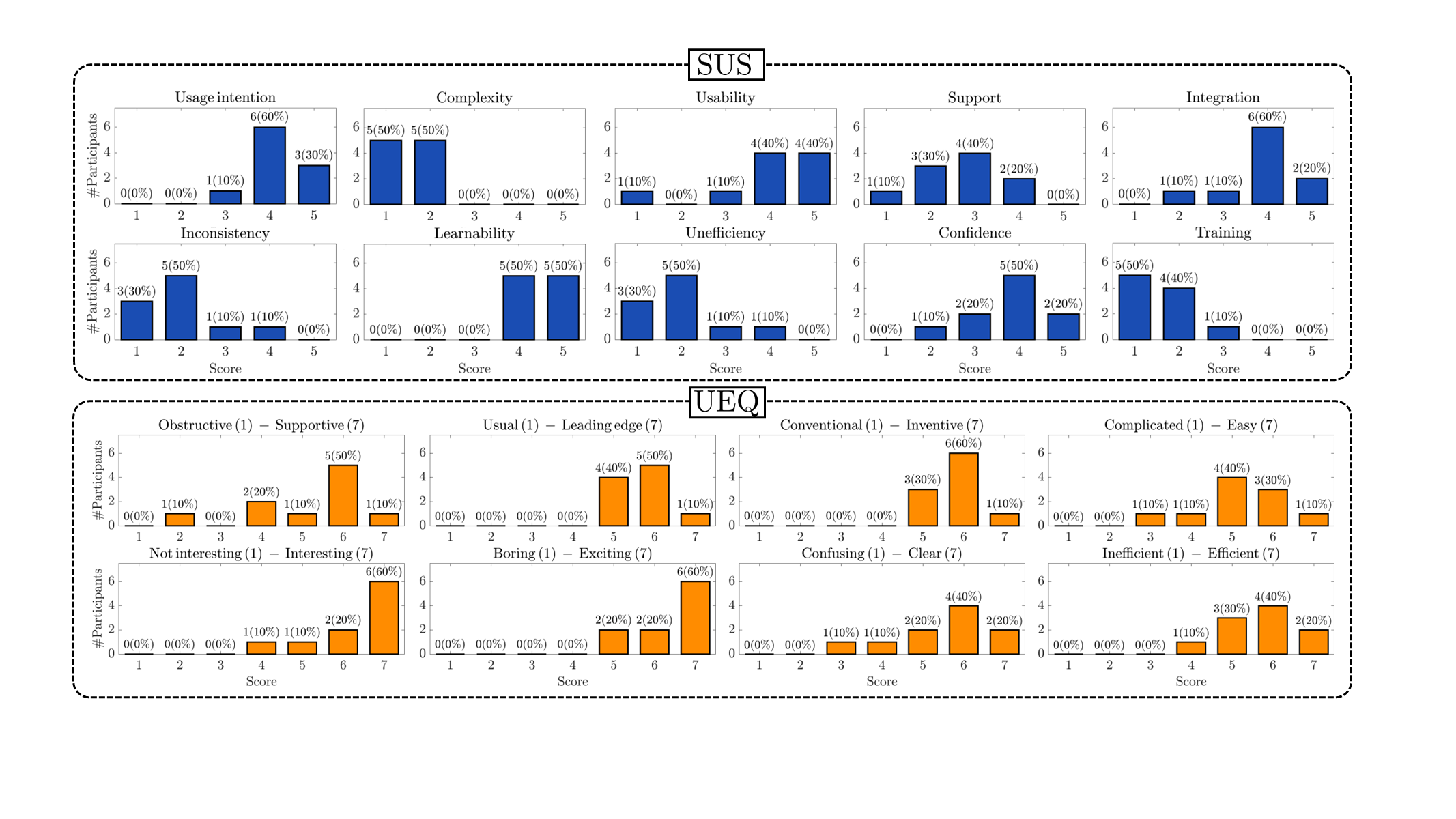}
    \caption{System usability scale (SUS) and user experience questionnaires (UEQ) outcomes for the proposed framework.}
    \label{fig:user_questionnaires}
    \vspace{-5mm}
\end{figure*}

To evaluate the proposed teleoperation system from a user perspective, participants were asked to complete the System Usability Scale (SUS) \cite{bangor2008empirical} and the User Experience Questionnaire (UEQ) \cite{laugwitz2008construction}. The SUS assesses perceived system usability, while the UEQ provides a structured evaluation of user experience across multiple dimensions. The resulting scores are reported in Fig.~\ref{fig:user_questionnaires}.

The usability and user experience evaluations indicate a generally positive perception of the proposed teleoperation system. SUS results suggest good overall usability, with participants reporting high usage intention, effective integration of system functions, and strong learnability, indicating that the system can be approached easily and becomes usable within a short time. Users also reported feeling comfortable and confident during interaction. At the same time, responses to negatively phrased items revealed perceived complexity, occasional inconsistency, inefficiency, and a need for further training, suggesting that cognitive and operational demands remain and could be mitigated through continued refinement and user-centered optimization. Consistently, UEQ results show a strongly positive overall user experience, with the system perceived as supportive, innovative, and engaging. High ratings related to clarity, efficiency, and ease of use indicate that interaction was well understood despite the system’s technical complexity, while scores reflecting novelty and inventiveness highlight its perceived technological advancement in the context of robotic-assisted TLM.

\section{Discussion \& Conclusion}\label{sec:Discussion}

In transoral microsurgery, tissue manipulation is typically performed manually using forceps, which may reduce precision and increase surgeon fatigue during prolonged procedures. This paper presented a novel teleoperation system for transoral microsurgery that integrates a computer-controlled forceps end-effector within a complete robotic teleoperation framework, enabling remote manipulation, forceps actuation, and enforcement of a RCM constraint.

Experimental results demonstrated improved grasping and manipulation stability compared to freehand operation, along with reduced muscular effort as indicated by sEMG analysis. In addition, usability assessments based on SUS and UEQ questionnaires showed a generally positive user perception of the proposed system.
Some limitations were observed, primarily related to the limited training time provided to participants, which occasionally affected task completion time. Participants also indicated that adding 3D visualization and the integration of force feedback could further improve system usability. Future work will therefore focus on incorporating haptic feedback, refining motion constraints to prevent laryngoscope occlusion, and investigating user learning effects over longer training sessions.



\bibliographystyle{IEEEtran}
\bibliography{references}

@ARTICLE{acemoglu2020,
  oldauthor={Acemoglu, Alperen and Krieglstein, Jan and Caldwell, Darwin G. and Mora, Francesco and Guastini, Luca and Trimarchi, Matteo and Vinciguerra, Alessandro and Carobbio, Andrea Luigi Camillo and Hysenbelli, Juljana and Delsanto, Marco and Barboni, Ottavia and Baggioni, Sabrina and Peretti, Giorgio and Mattos, Leonardo S.},
  author={Acemoglu, Alperen and others},
  journal={IEEE Trans. Med. Robot. Bionics}, 
  title={5G Robotic Telesurgery: Remote Transoral Laser Microsurgeries on a Cadaver}, 
  year={2020},
  volume={2},
  number={4},
  pages={511-518},
  }

@article{mohan2021telesurgery,
  title={Telesurgery and robotics: an improved and efficient era},
  author={Mohan, Anmol and Wara, Um Ul and Shaikh, Muhammad Taha Arshad and Rahman, Rahil M and Zaidi, Zain Ali and Shaikh, Muhammad Taha A},
  journal={Cureus},
  volume={13},
  number={3},
  year={2021},
  publisher={Cureus}
}

@article{patel2022haptic,
  title={Haptic feedback and force-based teleoperation in surgical robotics},
  author={Patel, Rajni V and Atashzar, S Farokh and Tavakoli, Mahdi},
  journal={Proceedings of the IEEE},
  volume={110},
  number={7},
  pages={1012--1027},
  year={2022},
  publisher={IEEE}
}

@article{barba2022remote,
  title={Remote telesurgery in humans: a systematic review},
  author={Barba, Patrick and Stramiello, Joshua and Funk, Emily K and Richter, Florian and Yip, Michael C and Orosco, Ryan K},
  journal={Surgical endoscopy},
  volume={36},
  number={5},
  pages={2771--2777},
  year={2022},
  publisher={Springer}
}

@article{jiang2022deep,
  title={Deep learning techniques for tumor segmentation: a review},
  author={Jiang, Huiyan and Diao, Zhaoshuo and Yao, Yu-Dong},
  journal={The Journal of Supercomputing},
  volume={78},
  number={2},
  pages={1807--1851},
  year={2022},
  publisher={Springer}
}

@inproceedings{petersen2025development,
  title={Development of a Bioimpedance Probe for Enhanced Tissue Identification in the Upper Aerodigestive Tract},
  author={Petersen, Marianne Viola and Savarimuthu, Thiusius Rajeeth and Cheng, Zhuoqi},
  booktitle={Proc. Annu. Int. Conf. IEEE Eng. Med. Biol. Soc. (EMBC)},
  volume={2025},
  pages={1--6},
  year={2025}
}

@article{pacheco2024automatic,
  title={Automatic Focus Adjustment for Single-Spot Tissue Temperature Control in Robotic Laser Surgery},
  author={Pacheco, Nicholas E and Gaddipati, Chaitanya S and Farzan, Siavash and Fichera, Loris},
  journal={IEEE Trans. Med. Robot. Bionics},
  year={2024},
  publisher={IEEE}
}

@article{carrieri2025design,
  title={Design and Control of a Computer-Assisted Laser Microsurgery System Based on Rotary Voice Coil Actuators},
  author={Carrieri, Lapo and Di Lucchio, Michele and Giannoni, Luca and Acemoglu, Alperen and Mattos, Leonardo S},
  journal={IEEE Trans. Med. Robot. Bionics},
  year={2025},
  publisher={IEEE}
}

@article{ansarin2014transoral,
  title={Transoral robotic surgery vs transoral laser microsurgery for resection of supraglottic cancer: a pilot surgery},
  oldauthor={Ansarin, Mohssen and Zorzi, Stefano and Massaro, Maria Angela and Tagliabue, Marta and Proh, Michele and Giugliano, Gioacchino and Calabrese, Luca and Chiesa, Fausto},
  author={Ansarin, Mohssen and others},
  journal={Int. J. Med. Robot. Comput. Assist. Surg.},
  volume={10},
  number={1},
  pages={107--112},
  year={2014},
  publisher={Wiley Online Library}
}

@article{o2010methodology,
  title={A methodology for design and appraisal of surgical robotic systems},
  author={O'toole, Michael d and Bouazza-Marouf, Kaddour and Kerr, David and Gooroochurn, Mahendra and Vloeberghs, Michael},
  journal={Robotica},
  volume={28},
  number={2},
  pages={297--310},
  year={2010},
  publisher={Cambridge University Press}
}

@article{lee2022end,
  title={When the end effector is a laser: A review of robotics in laser surgery},
  author={Lee, Hun Chan and Pacheco, Nicholas E and Fichera, Loris and Russo, Sheila},
  journal={Advanced Intelligent Systems},
  volume={4},
  number={10},
  pages={2200130},
  year={2022},
  publisher={Wiley Online Library}
}

@article{chauhan2019robotic,
  title={A robotic microsurgical forceps for transoral laser microsurgery},
  oldauthor={Chauhan, Manish and Deshpande, Nikhil and Pacchierotti, Claudio and Meli, Leonardo and Prattichizzo, Domenico and Caldwell, Darwin G and Mattos, Leonardo S},
  author={Chauhan, Manish and others},
  journal={International journal of computer assisted radiology and surgery},
  volume={14},
  number={2},
  pages={321--333},
  year={2019},
  publisher={Springer}
}

@inproceedings{deshpande2016robot,
  title={Robot-assisted microsurgical forceps with haptic feedback for transoral laser microsurgery},
  author={Deshpande, Nikhil and Chauhan, Manish and Pacchierotti, Claudio and Prattichizzo, Domenico and Caldwell, Darwin G and Mattos, Leonardo S},
  booktitle={Proc. 38th Annu. Int. Conf. IEEE Eng. Med. Biol. Soc. (EMBC)},
  pages={5156--5159},
  year={2016},
  organization={IEEE}
}

@article{bangor2008empirical,
  title={An empirical evaluation of the system usability scale},
  author={Bangor, Aaron and Kortum, Philip T and Miller, James T},
  journal={Intl. Journal of Human--Computer Interaction},
  volume={24},
  number={6},
  pages={574--594},
  year={2008},
  publisher={Taylor \& Francis}
}

@inproceedings{laugwitz2008construction,
  title={Construction and evaluation of a user experience questionnaire},
  author={Laugwitz, Bettina and Held, Theo and Schrepp, Martin},
  booktitle={Symposium of the Austrian HCI and usability engineering group},
  pages={63--76},
  year={2008},
  organization={Springer}
}

@inproceedings{ebied2020upper,
  title={Upper limb muscle fatigue analysis using multi-channel surface EMG},
  author={Ebied, Ahmed and Awadallah, Ahmed M and Abbass, Mohamed A and El-Sharkawy, Yasser},
  booktitle={2020 2nd Novel Intelligent and Leading Emerging Sciences Conference (NILES)},
  pages={423--427},
  year={2020},
  organization={IEEE}
}

@article{hou2021effects,
  title={Effects of various physical interventions on reducing neuromuscular fatigue assessed by electromyography: a systematic review and meta-analysis},
  author={Hou, Xiao and Liu, Jingmin and Weng, Kaixiang and Griffin, Lisa and Rice, Laura A and Jan, Yih-Kuen},
  journal={Frontiers in bioengineering and biotechnology},
  volume={9},
  pages={659138},
  year={2021},
  publisher={Frontiers Media SA}
}

@inproceedings{aghakhani2013task,
  title={Task control with remote center of motion constraint for minimally invasive robotic surgery},
  author={Aghakhani, Nastaran and Geravand, Milad and Shahriari, Navid and Vendittelli, Marilena and Oriolo, Giuseppe},
  booktitle={2013 IEEE international conference on robotics and automation},
  pages={5807--5812},
  year={2013},
  organization={IEEE}
}

@article{nasiri2024teleoperation,
  title={Teleoperation in robot-assisted MIS with adaptive RCM via admittance control: E. Nasiri et al.},
  author={Nasiri, Ehsan and Sowrirajan, Srikarran and Wang, Long},
  journal={International Journal of Intelligent Robotics and Applications},
  volume={8},
  number={4},
  pages={827--839},
  year={2024},
  publisher={Springer}
}

@article{colan2020cooperative,
  title={A cooperative human-robot interface for constrained manipulation in robot-assisted endonasal surgery},
  author={Colan, Jacinto and Nakanishi, Jun and Aoyama, Tadayoshi and Hasegawa, Yasuhisa},
  journal={Applied Sciences},
  volume={10},
  number={14},
  pages={4809},
  year={2020},
  publisher={MDPI}
}

\end{document}